\pdfoutput=1
\PassOptionsToPackage{table}{xcolor}
\documentclass{bmvc2k}

\newcommand{\paperTitleShort}{M\textsuperscript{3}T}
\newcommand{\paperTitleOne}{\paperTitleShort: Discrete Multi-Modal Motion Tokens for Sign Language Production}
\newcommand{\projectPage}{https://cogvis-cvssp.github.io/papers/m3t/}

\usepackage{amsmath}
\usepackage{microtype}
\usepackage{caption}
\usepackage{colortbl}
\usepackage[hang,flushmargin]{footmisc}

\newcommand{\emphif}[1]{{\emph{#1}}}

\usepackage[nolist]{acronym}
\usepackage{amssymb}
\usepackage{multirow}
\usepackage{tabularx}
\usepackage{booktabs}
\usepackage{adjustbox}
\usepackage{pifont}
\usepackage{bm}
\usepackage{soul}
\usepackage{subcaption}
\usepackage[capitalise]{cleveref}

\definecolor{softpink}{RGB}{220, 65, 130}

\title{\paperTitleOne}

\addauthor{Alexandre Symeonidis-Herzig}{a.symeonidisherzig@surrey.ac.uk}{1}
\addauthor{Jianhe Low}{jianhe.low@surrey.ac.uk}{1}
\addauthor{Ozge Mercanoglu Sincan}{o.mercanoglusincan@surrey.ac.uk}{1}
\addauthor{Richard Bowden}{r.bowden@surrey.ac.uk}{1}

\addinstitution{
 CVSSP\\
 University of Surrey\\
 United Kingdom
}

\runninghead{A. Symeonidis-Herzig et al.}{\paperTitleShort: Multi-Modal Motion Tokens for SLP}

\begin{document}

\maketitle

\begin{acronym}

\acro{3dgs}[3DGS]{3D Gaussian Splatting}
\acro{3dmm}[3DMM]{3D Morphable Model}
\acro{bsl}[BSL]{British Sign Language}
\acro{dgs}[DGS]{German Sign Language}
\acro{asl}[ASL]{American Sign Language}
\acro{csl}[CSL]{Chinese Sign Language}
\acro{slr}[SLR]{Sign Language Recognition}
\acro{cnn}[CNN]{Convolutional Neural Network}
\acro{lbs}[LBS]{Linear Blend Skinning}
\acro{vae}[VAE]{Variational Autoencoder}
\acro{ste}[STE]{straight-through estimator}
\acro{smlm}[SMLM]{Sign Motion Language Model}

\acro{nlp}[NLP]{Natural Language Processing}

\acro{gan}[GAN]{Generative Adversarial Network}
\acro{gpu}[GPU]{graphics processing unit}

\acro{lm}[LM]{Language Model}
\acro{llm}[LLM]{Large Language Model}

\acro{ldm}[LDM]{Latent Diffusion Model}

\acro{nmf}[NMF]{Non-Manual Feature}

\acro{sl}[SL]{sign language}
\acro{slp}[SLP]{Sign Language Production}
\acro{slt}[SLT]{Sign Language Translation}

\acro{sota}[SOTA]{state-of-the-art}
\acro{vq}[VQ]{Vector Quantized}
\acro{fsq}[FSQ]{Finite Scalar Quantized}

\acro{bt}[BT]{Back-Translation}

\end{acronym}

\begin{abstract}
Sign language production requires more than hand motion generation.
Non-manual features, including mouthings, eyebrow raises, gaze, and head movements, are grammatically obligatory and cannot be recovered from manual articulators alone.
Existing 3D production systems face two barriers to integrating them: the body-model fittings they train on retain a facial space too low-dimensional to encode these articulations, and, if richer representations are adopted, standard discrete tokenization suffers from codebook collapse, leaving most of the expression space largely unreachable.
We propose SMPL-FX, which couples FLAME's rich expression space with the SMPL-X body parameterisation, and tokenize the resulting representation with modality-specific Finite Scalar Quantization VAEs for body, hands, and face, raising face codebook utilization from 78.8\% to 99.0\%.
\paperTitleShort{} is an autoregressive transformer trained on this multi-modal motion vocabulary, with an auxiliary sign-to-text translation objective that encourages semantically grounded embeddings.
Across three standard datasets, \paperTitleShort{} achieves state-of-the-art sign language production quality, and on NMFs-CSL, where signs are distinguishable only by non-manual features, our model reaches 58.3\% accuracy against 49.0\% for the strongest comparable baseline, without large-scale sign-language pre-training.

\noindent Project page: \url{\projectPage}
\end{abstract}

\section{Introduction}
\label{sec:intro}
Producing sign language requires the coordinated generation of two articulatory channels: the manual channel, comprising hand shape, orientation, and motion, and the non-manual channel, comprising facial expression, mouthing, gaze, and head dynamics~\cite{stokoe1980sign,sutton1999linguistics,Hands_Are_The_Head_of_The_Mouth}.
Critically, \acp{nmf} are grammatically obligatory: they negate statements, alter grammatical mood, or distinguish between otherwise identical manual signs~\cite{stokoe1980sign}.
Generating these cues faithfully is therefore essential for \ac{slp} systems to achieve accurate and grammatically complete signing.

Despite their importance, current 3D \ac{slp} systems focus primarily on manual features or rely on low-dimensional facial models.
This renders them unable to distinguish signs that share identical manual form and are differentiated solely by mouthing~\cite{sutton1999linguistics}.
For example, in \ac{csl} the signs for \emph{bike} and \emph{hesitate} share identical manual articulation and are disambiguated only by their accompanying \acp{nmf}; a system that cannot generate these features collapses such pairs into a single output.

Two limitations in current 3D \ac{slp} pipelines underlie this gap.
First, although SMPL-X can expose FLAME's full expression space, the whole-body fittings adopted by prior tokenized systems recover only ten expression coefficients, insufficient to represent the articulations \acp{nmf} require.
Second, VQ-VAE tokenization of low-dimensional PCA-based face parameters suffers from codebook collapse, yielding 78.8\% face codebook utilization in our experiments.
This collapse is more consequential for the face: unlike body and hand parameters, which are joint rotations, expression PCA coefficients form a compact, dense space in which missing codebook entries directly limit which expressions can be reconstructed. The closest prior work, SOKE~\cite{zuo2025soke}, shares both limitations.

We adopt the discrete motion-language-modelling paradigm~\cite{jiang2023motiongpt,wang2024motiongpt2} for \ac{slp} and address these limitations with three contributions.
First, we extract SMPL-FX fittings that combine the SMPL-X body with FLAME's 100-dimensional expression head~\cite{FLAME,pavlakos2019expressive}, providing the facial resolution that prior 3D sign datasets lack; we release these fittings as a resource for sign language research.
Second, modality-specific FSQ-VAEs~\cite{mentzer2022finite} raise face codebook utilization to 99\%, recovering the expressions that VQ omits and improving overall reconstruction quality.
Third, our autoregressive transformer is trained over the resulting multi-modal token vocabulary with an auxiliary \ac{slt} objective which further improves geometric fidelity and back-translation quality across all benchmarks.

Across three sign language datasets (Phoenix14T~\cite{Camgoz_2018_CVPR_pheonix14}, CSL-Daily~\cite{zhou2021improvingcsldaily}, How2Sign~\cite{duarte2021how2sign}), \paperTitleShort\ achieves state-of-the-art production quality without retrieval from external sign dictionaries.
On the NMFs-CSL benchmark~\cite{hu2021global}, which contains signs distinguishable only by non-manual facial features, \paperTitleShort\ achieves 58.3\% Top-1 accuracy against 49.0\% for the leading pose-based baseline without large-scale pretraining, showing that the learned tokens capture the grammatical distinctions that prior systems represent too coarsely to recover.

\section{Related Work}\label{sec:related}

\noindent\textbf{Human Motion Modeling.}
\emphif{Motion generation}, synthesizing body motion conditioned on audio, prior motion, or text~\cite{zhang2023t2mgpt,jiang2023motiongpt,zuo2025soke}, is a long-standing problem in graphics and vision~\cite{badler1993simulating,kojima2002natural}.
Recent text-to-motion methods fall into two dominant architectures: diffusion models~\cite{latentdiff,ho2020denoising,song2020score,zhang2024motiondiffuse,chen2023mld,guy2022humanmdm}, which generate continuous trajectories in latent space, and tokenized autoregressive models~\cite{zhang2023t2mgpt,jiang2023motiongpt,zuo2025soke}, which discretize motion via codebooks and train \ac{llm} backbones~\cite{liu2020multilingual,raffel2020exploringt5}.
While diffusion emphasizes motion realism, tokenized formulations cast motion as a symbolic vocabulary that a language model can directly process for both text-conditioned generation and for \emphif{motion captioning}~\cite{plappert2018learning,takano2015statistical}, the inverse task of describing motion in natural language.
Prior works have trained these tasks together in shared architectures~\cite{zhang2023t2mgpt,jiang2023motiongpt}; while their primary goal is task unification, the shared vocabulary may also act as a semantic anchor, encouraging tokens to retain linguistically relevant features within the language model's representations.
Structured human generation under explicit conditional control has also been studied on the appearance side, e.g.\ pose-guided person image generation~\cite{shen2024imagpose} and customizable virtual dressing~\cite{shen2025imagdressing}, complementing the motion-centric control considered here.

\noindent\textbf{Sign Language Translation and Production.}
\ac{slp} (text-to-sign) and \ac{slt} (sign-to-text) are conceptually symmetric, yet largely developed in isolation~\cite{chaudhary2022signnet}.
SLP has progressed from rule-based and retrieval systems~\cite{glauert2006vanessa,karpouzis2007educational,mcdonald2016automated} to neural two-stage pipelines~\cite{saunders2020everybody,stoll2020text2sign}, and more recently transformer-~\cite{saunders2020adversarial,huang2021towards} and diffusion-based models~\cite{baltatzis2024neural}.
Across both 3D body-model pipelines and video-based rendering approaches, the incorporation of facial \acp{nmf} remains a challenge, with facial articulation largely absent from current systems~\cite{qi2024signgen,fang2025signdiff,wang2025advancedsignlanguagevideo} regardless of the underlying representation.
In 3D pose-based pipelines, a common shortcoming lies in the fittings rather than the body model itself: although SMPL-X~\cite{pavlakos2019expressive} supports FLAME's full expression space, the whole-body estimators used by prior \ac{slp} pipelines~\cite{zuo2025soke,baltatzis2024neural} regress only ten expression coefficients, insufficient to encode the mouthings and facial expressions required in sign.
Conversely, SLT models map video or pose sequences to text using gloss\footnote{A gloss is a written approximation of a sign's meaning.}-based supervision~\cite{Camgoz_2018_CVPR_pheonix14,hu2021global} or end-to-end transformers~\cite{vaswani2017attention,camgoz2020sign}, increasingly strengthened by pretrained \acp{llm}~\cite{chen2022simple,chentwo,jiao2024visual,Wei_2023_ICCV}.

While emerging SLT methods~\cite{jang2025losttranslationembeddingssign,li2025unisign} have begun incorporating facial cues, \ac{nmf} integration in production remains underexplored, despite their inclusion being mandatory for sign.

\noindent\textbf{Discrete Motion Representation.}
Discrete quantization underpins modern generative modeling, from images~\cite{oord2018neural,esser2021taming} to motion~\cite{zhang2023t2mgpt,jiang2023motiongpt,wang2024motiongpt2,chuan2022tm2t}.
VQ-\acp{vae}~\cite{oord2018neural,razavi2019generating} replace continuous latents with nearest codebook entries, enabling transformer-based motion generation.
Sign language production has adopted VQ tokenization~\cite{walsh2024data,zuo2025soke}, but primarily for body and hand motion.
When applied to high-dimensional facial parameters, VQ-VAEs suffer codebook collapse: in our experiments VQ achieves only 78.8\% face codebook utilization, leaving large regions of the expression space unreachable.
\ac{fsq}~\cite{mentzer2022finite} removes the learned codebook entirely, partitioning the latent space into a structured grid, leading to improved utilization and reconstruction.
SignViP~\cite{wang2025advancedsignlanguagevideo} also applies FSQ in sign, but quantizes 2D pose as conditioning for a video diffusion generator; our FSQ tokens are the 3D motion vocabulary itself, modelled autoregressively over SMPL-FX body, hands, and face.
Extending FSQ to the face modality, where codebook collapse is most severe, is the step required for discrete facial tokens to carry \acp{nmf} present in sign.

\section{Method}\label{sec:method}

\begin{figure*}[htbp]
  \centering
  \includegraphics[trim={0.10cm 0.20cm 0.10cm 0.1cm},clip,width=\linewidth]{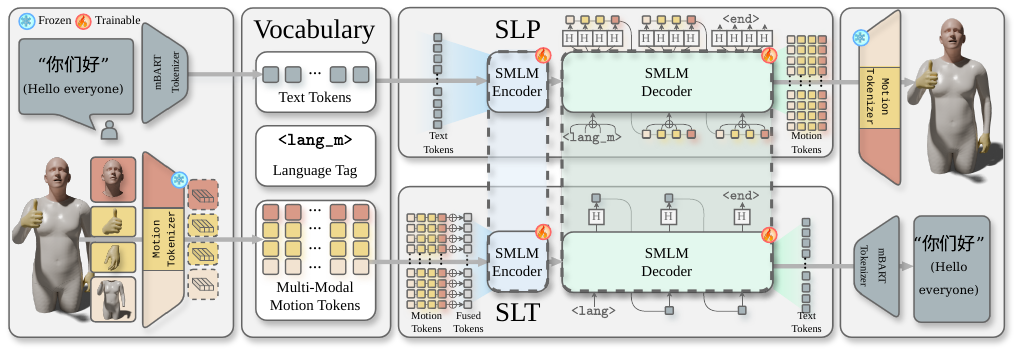}
  \vspace{6pt}
    \caption{\textbf{Sign Motion Language Model.}
    A single autoregressive transformer operates over \ac{slp} (\texttt{text}$\rightarrow$\texttt{motion})
    and \ac{slt} (\texttt{motion}$\rightarrow$\texttt{text}) within a shared token space.
    Spoken-language tokens and modality-specific FSQ motion tokens (body, hands, face)
    are embedded and concatenated with identifiers
    (e.g., \texttt{<ASL\_RH>}, \texttt{<DGS\_F>}) to distinguish language and articulator streams.
    The model predicts the next token across all modalities using a unified encoder and decoder with modality-specific output heads.
    The translation supervision grounds token embeddings, improving production quality.
    Both production and translation share the same encoder-decoder parameters.}\label{fig:mlm}
\end{figure*}

Multi-Modal Motion Tokens (\paperTitleShort) is a sign language production framework (\cref{fig:mlm}) built on rich discrete motion representations spanning the face, body, and hands.
\paperTitleShort{} comprises two components: (1) modality-specific \ac{fsq}-\ac{vae} tokenizers that convert continuous SMPL-FX parameters into temporally aligned streams of discrete tokens, and (2) our \ac{smlm}, an autoregressive transformer that is trained on these tokens for \ac{slp} and is regularized by an auxiliary translation objective.

We address the previously identified limitations with SMPL-FX (\cref{sec:3dmm_prep}) and per-modality FSQ tokenizers (\cref{sec:tokenization}).
Additionally, a \ac{slr} module (\cref{sec:recognition}) probes whether the resulting tokens carry the linguistic distinctions, particularly \acp{nmf}.
We begin with the SMPL-FX reconstruction pipeline.

\subsection{Human Reconstruction} \label{sec:3dmm_prep}
The first stage of our pipeline extracts temporally coherent \ac{3dmm} parameters from monocular signing footage, providing the representation on which all downstream tokenization and generation operates.
Recent monocular reconstruction advances make this tractable at scale~\cite{potamias2025wilor,sarandi2024neural_nlf,giebenhain2025pixel3dmm}, and large-scale extraction pipelines across continuous sign corpora have been demonstrated~\cite{baltatzis2024neural,zuo2025soke,duarte2021how2sign,zhou2021improvingcsldaily,Camgoz_2018_CVPR_pheonix14}.
Sign language places a specific demand on facial representation that general body estimation approaches do not address.
Many signs share identical manual form, i.e., the same hand shape, trajectory, and location, but carry entirely distinct meanings solely through the word being mouthed~\cite{Hands_Are_The_Head_of_The_Mouth}.
Recovering \acp{nmf} such as these requires accurate facial reconstruction beyond the resolution of standard SMPL-X reconstructions.

\begin{figure}[htbp]
    \centering
    \begin{minipage}[c]{0.48\textwidth}
        \centering
        \includegraphics[width=\linewidth]{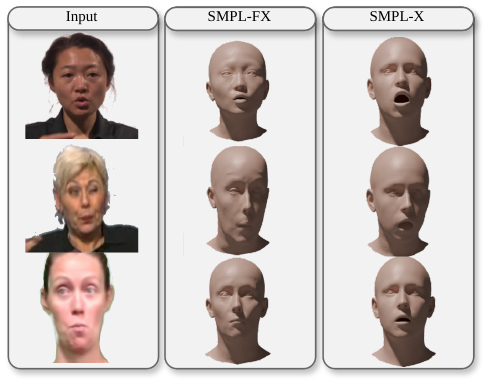}
    \end{minipage}
    \hfill
    \begin{minipage}[c]{0.48\textwidth}
        \centering
        \includegraphics[width=\linewidth]{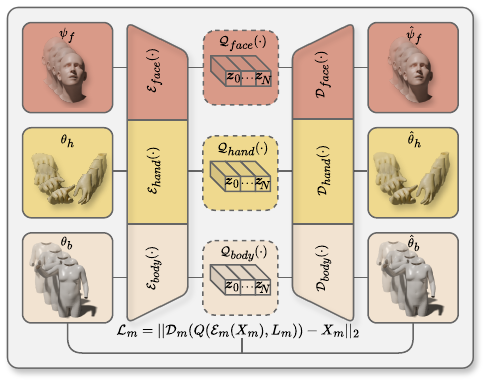}
    \end{minipage}

    \vspace{0.4cm}

    \begin{minipage}[t]{0.48\textwidth}
        \caption{\textbf{SMPL-FX (ours) vs.\ SMPL-X.} Facial parameters for SMPL-FX are extracted via Pixel3DMM~\cite{giebenhain2025pixel3dmm}; for SMPL-X we use parameters as extracted by SOKE~\cite{zuo2025soke}. The SMPL-X captures primarily head rotation, failing to capture mouth shape and any smaller expressions. This compares full reconstruction pipelines rather than isolated parameterisations; for production, this is the comparison that matters, as a model cannot learn expressions its training fittings never contain.}
        \label{fig:smplfx}
    \end{minipage}
    \hfill
    \begin{minipage}[t]{0.48\textwidth}
        \caption{\textbf{Multi-modal FSQ-VAE Tokenization.} Our FSQ-VAE framework discretizes modality-specific SMPL-FX latents into a structured grid. The resulting token streams provide a unified, compact, and discrete multi-modal representation specifically optimized for signing motion.}
        \label{fig:token}
    \end{minipage}
\end{figure}

\cref{fig:smplfx} shows SMPL-X~\cite{pavlakos2019expressive} fittings used by prior 3D \ac{slp} systems~\cite{zuo2025soke} and the resulting inaccurate faces.
While head pose is reasonably well captured, the facial parameters collapse to near-constant values across the sequence, resulting in a static face that cannot encode the \acp{nmf} required for accurate signing.
This is a limitation of the fittings rather than of SMPL-X itself: the model supports FLAME's full expression space, but whole-body estimators typically regress only ten coefficients, and recovering the remainder by optimization on monocular video is under-constrained. Dedicated FLAME estimators such as Pixel3DMM~\cite{giebenhain2025pixel3dmm} instead predict the full expression space feed-forward.
All models trained on these parameters have no access to these distinctions and therefore cannot produce linguistically complete signing.

To alleviate this, we construct \emphif{SMPL-FX}: an upper-body model that replaces SMPL-X's head with the native FLAME~\cite{FLAME} mesh and its 100-dimensional expression space, allowing such estimators to drive the face directly and providing the resolution needed to capture mouthings and the full range of grammatically relevant \acp{nmf}.
As shown in \cref{fig:smplfx}, SMPL-FX both captures these details and provides a more accurate head shape, useful for downstream rendering and avatar production.
Since the lower body contributes minimally to signing, we retain only the torso and arms, removing 11 irrelevant joints for efficiency.
Each frame of signing motion is encoded as:
\begin{equation}\label{eq:smplfx_in}
    \boldsymbol{\theta}_{h}^{l},\boldsymbol{\theta}_{h}^{r}\in\mathbb{R}^{15\times6},\;
    \boldsymbol{\theta}_{h}=\{\boldsymbol{\theta}_{h}^{l},\boldsymbol{\theta}_{h}^{r}\},\;
    \boldsymbol{\theta}_{b}\in\mathbb{R}^{10\times6},\;
    \boldsymbol{\psi}_{f}\in\mathbb{R}^{108},
\end{equation}
denoting hand, body, and face parameters; a static identity blendshape $\beta$ is also estimated per signer (see \cref{supp:smplfx} for the parameter breakdown).
Because $\boldsymbol{\theta}_{h}$ and $\boldsymbol{\theta}_{b}$ are initially estimated per frame, we further refine them across the sequence to regularize joint velocity and minimize 2D keypoint reprojection error~\cite{yang2023effective_dwpose}; the full per-sequence refinement and quality-control procedure is detailed in \cref{supp:smplfx}.
With optimized parameters, generation of the full mesh is then performed as:
\begin{equation}\label{eq:smplfx}
    (\mathbf{V}, \mathbf{J}) = \mathrm{SMPL\text{-}FX}(\beta, \boldsymbol{\theta}_{h}, \boldsymbol{\theta}_{b}, \boldsymbol{\psi}_{f}),
\end{equation}
where $\mathbf{V}\!\in\!\mathbb{R}^{N_v\times3}$ ($N_v=10{,}995$) are vertex coordinates and $\mathbf{J}\!\in\!\mathbb{R}^{N_j\times3}$ ($N_j=42$) are 3D joint locations.
Operating directly in the \ac{3dmm} domain, as opposed to pose, also facilitates real-time rendering of photorealistic, sign-capable avatars~\cite{zhang2025guavageneralizableupperbody,ivashechkin2025signsplat,walsh2025using} through \ac{3dgs} pipelines.
SMPL-FX further augments the face with rigidly-attached teeth geometry~\cite{qian2024gaussianavatars} (not reflected in joint metrics, but required to render mouthings) and independent eyelid controls~\cite{giebenhain2025pixel3dmm} absent from SMPL-X.
To avoid neck artifacts at the FLAME--SMPL boundary, vertex positions are interpolated between SMPL-X and FLAME blend weights, weighting toward the body at the neck and toward the face near the jaw. Global alignment follows~\cite{zhang2025guavageneralizableupperbody}, matching eye displacement between meshes.
The value of SMPL-FX thus lies less in the coupling itself than in what it enables: to our knowledge, the resulting corpora fittings are the first for continuous sign data with \ac{nmf}-level facial detail.

\subsection{Multi-Modal Sign Motion Tokenization}\label{sec:tokenization}
Continuous SMPL-FX parameters are high-dimensional and not directly suited to sequence modeling.
We discretize each articulator stream into its own vocabulary using modality-specific FSQ-VAEs~\cite{mentzer2022finite} (\cref{fig:token}), producing token streams for the \ac{smlm}.

\noindent\textbf{Encoders:}
For each modality $m \in \{b,h,f\}$, an encoder $\mathcal{E}_m$ maps the \ac{3dmm} sequence $X_m \in \mathbb{R}^{T \times D_m}$ of $T$ frames and dimension $D_m$ (\cref{eq:smplfx_in}) to downsampled latents using a stack of 1D convolutional residual blocks~\cite{jiang2023motiongpt}, with window size $w$ and latent dimension $d$:
\begin{equation}\label{eq:enc}
Z_m = \mathcal{E}_m(X_m) \in \mathbb{R}^{\frac{T}{w} \times d}.
\end{equation}
Left hand poses are mirrored before encoding, enabling a shared encoder for the hands.

\noindent\textbf{Quantization:}
Applying VQ-VAE to the facial stream $\boldsymbol{\psi}_f \in \mathbb{R}^{108}$ produces codebook collapse, particularly as faces are parameterized via low-rank PCA coefficients rather than geometric pose angles.
As discussed in later sections (\cref{sec:ablation_studies}), our experiments reveal that VQ-VAE reaches only 78.8\% codebook utilization, leaving entire manifolds of the expression space unreachable and rendering discrete facial tokenization representationally bottlenecked.
FSQ~\cite{mentzer2022finite} resolves this with an element-wise quantization operator $Q(\cdot, L_m)$ that independently maps each dimension $i$ of $Z_m$ into a predefined level $[0, L_m^i - 1]$ ($L_m\in\mathbb{Z}^d$), where $\mathrm{rnd}_{\mathrm{ste}}$ denotes straight-through estimation~\cite{bengio2013estimating_ste} rounding to allow gradient flow:
\begin{equation}\label{eq:fsq}
  \hat{Z}_m=Q(Z_m,L_m)=
  \mathrm{rnd}_{\mathrm{ste}}
  \!\left(
    \frac{L_m}{2}\tanh(Z_m)
  \right).
\end{equation}
Unlike VQ-VAE~\cite{oord2018neural}, FSQ does not require a learned codebook or a commitment loss, eliminating the risk of codebook collapse and improving utilization of the discrete space.
Deterministic quantization shifts the representational burden onto the encoder and decoder, enabling highly compact latent dimensionalities while maintaining high-fidelity reconstruction.
Quantization enables us to represent each window of $w$ frames as a single discrete integer index selected from an implicit codebook of size $C_m=\prod_{i=1}^{d} L_m^i$.
This yields $\hat{Z}_m \in \{1,...,C_m\}^{\frac{T}{w}}$ as a concise, modality-aligned sequence of motion tokens.

\noindent\textbf{Decoder:}
The tokens $\hat{Z}_m$ can then be decoded using modality-specific decoders $\mathcal{D}_m(\cdot)$, the structure of which mirrors that of the encoders, to reconstruct the \ac{3dmm} parameters.
\begin{equation}\label{eq:dec}
\hat{X}_m = \mathcal{D}_m(\hat{Z}_m).
\end{equation}
As we employ FSQ, training seeks to minimize the $\ell_2$ reconstruction error across modalities without the need for auxiliary codebook losses or commitment terms:
\begin{equation}\label{eq:recon_loss}
    \mathcal{L}_{m} = ||\mathcal{D}_m(Q(\mathcal{E}_m(X_m), L_m)) - X_m||_2.
\end{equation}
Our tokenization additionally differs from prior \ac{slp} methods~\cite{zuo2025soke}.
We train each modality's FSQ-VAE in a separate run rather than jointly, so the body, hand, and face codebooks can specialise to their own dynamics, producing materially stronger tokens per channel.
We share the hand encoder between left and right via mirroring, doubling the hand training data and removing the handedness bias of per-hand encoders.
Together these give substantially higher utilization and lower reconstruction error.
The four token streams (body, left hand, right hand, face) make up the multi-modal vocabulary used by the \ac{smlm}.

\subsection{Sign Motion Language Model}\label{sec:sign_language_model}
Using these multi-modal motion FSQ-VAEs, we project motion trajectories into a sequence of sign tokens in the same embedding space as a pre-trained language model~\cite{liu2020multilingual} text vocabulary $\mathcal{V}_t$.
The encoder--decoder architecture and decoding strategy follow prior tokenized \ac{slp}~\cite{zuo2025soke} by design: we do not focus on the fundamentals of the \ac{lm}, but on the multi-modal token vocabulary and the auxiliary translation objective. LM-side advances are left as future work.
We construct a sign vocabulary per modality by introducing a unique ``word'' for every FSQ index:
$\mathcal{V}_m=\{\texttt{<m\_}1\texttt{>},\dots,\texttt{<m\_}C_m\texttt{>}\}$.
As is standard for multi-lingual models~\cite{liu2020multilingual,zuo2025soke}, we define language tokens $\mathcal{V}_l=\{\texttt{<lang\_m>}\}$ for each combination of language and modality.
These tags serve as domain prompts, allowing the same model to handle multiple languages.
Explicitly decoupling the hands yields four sign modalities $m\in\{b,lh,rh,f\}$, which when combined with pre-existing text and new language-modality tokens, produce our language model's complete lexicon:
\begin{equation} \label{eq:vocab}
\mathcal{V} = \mathcal{V}_t \cup \mathcal{V}_l \cup \mathcal{V}_{b} \cup \mathcal{V}_{lh} \cup \mathcal{V}_{rh} \cup \mathcal{V}_{f}.
\end{equation}

\noindent\textbf{Sign Language Production:}
Given a sequence of spoken-language text $\mathbf{S}$, the unified multi-modal vocabulary allows the model to autoregressively generate a sequence of sign-motion tokens $\mathbf{Y} = \{y_u\}_{u=1}^{U}$, where each time step contains one token per modality.
The input text sequence is first embedded and encoded by the transformer encoder.
Its final hidden states condition the decoder, which predicts the next multi-modal token set:
\begin{equation}
\label{eq:slp_ar}
P(\mathbf{Y} \mid \mathbf{S}) = \prod_{u=1}^{U} P\big(y_{u} \mid y_{<u}\big).
\end{equation}
To adapt the decoder to multi-modal generation, we follow~\cite{zuo2025soke} and prepend modality-specific language prompts (e.g., \texttt{<ASL\_LH>}, \texttt{<DGS\_F>}) to $\mathbf{Y}$.
Because the decoder expects a single input token per time step, we fuse the embeddings of the current step's motion tokens via an equal-weight average.
Per-modality discrimination is deferred entirely to the output heads.

For output, the transformer's hidden state is routed through parallel modality-specific linear heads, where each head predicts only the logits for its modality's vocabulary using the decoder's last hidden state.

\noindent\textbf{Auxiliary Translation Objective:}
Prior work has observed that sign-to-text supervision encourages more semantically grounded sign representations~\cite{walsh2025using,tang2022gloss}.
We incorporate this as an auxiliary training objective to regularize the motion generation. Given the motion-token streams $\mathbf{Y}$, their embeddings are fused and passed to the encoder, and the decoder autoregressively predicts the spoken-language sequence $\mathbf{S}=\{s_{k}\}_{k=1}^{K}$:
\begin{equation}
\label{eq:slt_ar}
P(\textbf{S} \mid \textbf{Y}) = \prod_{k=1}^{K} P\big(s_{k} \mid s_{<k}\big)
\end{equation}
using a single shared output head over the shared spoken-language vocabulary.
The translation loss serves as a semantic regularizer that further organizes the motion-token embedding space according to linguistic content, encouraging the encoder to cluster tokens with underlying signs that share lexical meaning.
The resulting semantic grounding consistently improves both geometric accuracy and back-translation quality.

Both objectives use standard cross-entropy losses.
During training, we alternate them by randomly sampling ($p=0.5$) production or translation per batch.
In the multi-stream setting, generation terminates when any modality emits the \texttt{<EOS>} token, truncating all four streams to a common length; this prevents the streams from diverging in length at the cost of occasionally ending a slower modality early.
At inference time, we employ greedy decoding, selecting the most probable token at each step.
For motion outputs, predicted discrete tokens are mapped back to \ac{3dmm} parameters using the frozen FSQ-VAE decoders, allowing mesh generation via \cref{eq:smplfx}.
These meshes can be rendered into signing videos using \ac{3dgs}-based~\cite{kerbl20233dgaussiansplatting} avatar synthesis~\cite{qian2024gaussianavatars,ivashechkin2025signsplat,Symeonidis-Herzig_2025_ICCV}.

\subsection{Sign Language Recognition}\label{sec:recognition}

\Ac{slr} maps video clips of isolated lexical signs to discrete class labels.
In the context of \paperTitleShort, we do not treat \ac{slr} as an end goal, but rather as a diagnostic tool to evaluate the representational fidelity of our learned multi-modal motion tokens.
Standard continuous metrics, such as joint position error, cannot fully capture whether vital components of sign, both manual and non-manual, have been preserved.
By evaluating \ac{slr} on subsets distinguishable only by \acp{nmf}, we directly probe whether face tokens carry the linguistic distinctions.

To isolate the quality of the representation, our modality-specific FSQ-VAE encoders remain frozen during this phase.
Following~\cite{wong2025signrep}, we first isolate the active signing window for each sequence $\mathbf{Y}$ by applying thresholds on hand height and displacement relative to a neutral rest pose, trimming uninformative initial and final frames.
These frames are then processed by per-modality encoders and we extract the final high-dimensional continuous hidden states immediately preceding the \ac{fsq} quantization step.
These modality-specific feature sequences are interleaved to form a structured temporal sequence, which is fed into a transformer encoder to capture global temporal dependencies, followed by a linear MLP classification head that predicts the sign class.

The classifier is trained using cross-entropy with label smoothing. Because the underlying motion encoders are frozen, the downstream \ac{slr} accuracy acts as a direct proxy for representation quality.
High performance on confusing signs, where manual features are identical and only the face discriminates similar signs, indicates that SMPL-FX captures the expression distinctions required for \acp{nmf} and that our learned tokens preserve these facial dynamics.

\section{Experiments}\label{sec:experiments}

We investigate whether our multi-modal tokens constitute a motion representation capable of driving \ac{slp} via three experiments: production quality on standard continuous sign benchmarks (\cref{sec:slp_results}), representational fidelity of facial tokens on the NMFs-CSL isolated sign recognition benchmark (\cref{sec:slr_experiments}), and tokenization design choices via ablation (\cref{sec:ablation_studies}).
Together, these experiments test whether SMPL-FX and FSQ tokenization produce motion tokens that are both geometrically accurate and linguistically discriminative, particularly for the facial channel that prior methods leave largely static.

\subsection{Experimental Setup}\label{sec:experimental_setup}

\noindent\textbf{Datasets:}
We evaluate on three continuous sign datasets frequently used in both \ac{slp} and \ac{slt}:
Phoenix14T~\cite{Camgoz_2018_CVPR_pheonix14} (\ac{dgs}, 8K sentences),
CSL-Daily~\cite{zhou2021improvingcsldaily} (\ac{csl}, 20K sentences), and
How2Sign~\cite{duarte2021how2sign} (\ac{asl}, 30K sentences).

To directly measure generalization of the facial tokens, we further evaluate \ac{slr} performance on the specialized NMFs-CSL~\cite{hu2021global}.
This benchmark contains 32K isolated signing videos spanning 610 ``confusing'' signs that are distinguishable only through \ac{nmf}, along with 457 ``normal'' signs, making it a targeted test for the representational fidelity, particularly for \acp{nmf}, encoded within our multi-modal motion tokens.

\noindent\textbf{Training Overview:}
Each FSQ-VAE modality is trained independently, and the three trained models are subsequently frozen for all downstream tasks.
Codebook sizes are selected to fairly compare with prior works at $|C_b|=100$, $|C_h|=150$, $|C_f|=216$, with a dimension of $C=3$ for all modalities at a temporal downsampling factor of $4\times$.
This greatly reduced dimensionality is a common trait of FSQ, as learning is pushed from the learned high-dimensional codebook entries into the encoder and decoder.

The unified transformer backbone is based on mBART-large~\cite{liu2020multilingual}, adapted to jointly process text and our motion vocabularies using a shared encoder and decoder with a linear layer per output modality.
Recognition is trained separately using a smaller transformer encoder and an MLP.
Additional architectural specifications, hyperparameters, and implementation details for each model are provided in \cref{supp:imp}.

\noindent\textbf{Metrics:}
We evaluate sign language production using standard metrics from 3D human motion generation.
Geometric fidelity, body and hand accuracy are measured using mean joint position error (JPE), computed both before and after rigid alignment via \emph{Procrustes analysis}.
For generated sequences, the joints are temporally aligned with \emph{Dynamic Time Warping} (DTW)~\cite{salvador2007dtw} prior to error computation.
To capture semantic preservation, we perform back-translation~\cite{saunders2020progressive}, where generated motion is translated back into text using separate \ac{slt} models.
We report BLEU-4~\cite{papineni2002bleu} and ROUGE-L~\cite{lin2004rouge}, the standard translation metrics, to quantify linguistic consistency between the generated and reference sentences across all three benchmarks.
We emphasize that back-translation is used here as a diagnostic of token semantics rather than as a translation system, and results are centered on the gap between generated motion and the GT-3DMM upper bound (\cref{tab:slp_extra}), not the absolute BLEU-4; \ac{slp} has no standardized back-translation protocol, and recent work documents clear tradeoffs among pose-based sign evaluation metrics~\cite{jiang2025meaningful}.
For facial motion, we report vertex position error (VPE) under DTW alignment, computed over the FLAME mesh vertices.
As back-translation models differ across methods, we train a single common~\cite{huang2021towards,saunders2020progressive,saunders2021mixed, walsh2025using} transformer-based \ac{slt} model~\cite{camgoz2020sign} on ground-truth 3DMM parameters to provide a consistent evaluation baseline.
We additionally report an upper bound obtained by passing ground-truth 3DMM parameters through the back-translation models.

Direct geometric comparison of facial outputs across methods is not feasible, as each system extracts pseudo-GT via a distinct body model and reconstruction pipeline; we detail our face-channel evaluation strategy in \cref{sec:slp_results}.

\begin{table*}[htbp]
\small
\setlength\tabcolsep{4pt}
\centering
\captionsetup{skip=6pt}
\caption{\textbf{Comparison with state-of-the-art \ac{slp} (text-to-sign) methods}.
DTW-PA-JPE measures pose-aligned joint error after rigid normalization, while DTW-JPE measures raw joint position error.
All metrics are computed with dynamic time warping to account for temporal variation between generated and reference sequences; lower values indicate higher motion fidelity.
Metrics marked with ${\blacklozenge}$ and ${\Diamond}$ are reported from~\cite{baltatzis2024neural} and~\cite{zuo2025soke}, respectively. ${\dagger}$ is MotionGPT~\cite{jiang2023motiongpt} as trained by~\cite{zuo2025soke} per sign dataset. $^{\ddagger}$~SOKE uses retrieval-augmented generation with external isolated sign dictionaries; \paperTitleShort\ achieves superior results without external retrieval or lexicon supervision.}
\label{tab:sota_slp}
\resizebox{\linewidth}{!}{
\begin{tabular}{l cccc cccc cccc}
\toprule
\multirow{3}{*}{Method} &
\multicolumn{4}{c}{\textbf{How2Sign}} &
\multicolumn{4}{c}{\textbf{CSL-Daily}} &
\multicolumn{4}{c}{\textbf{Phoenix14T}} \\
& \multicolumn{2}{c}{DTW-PA-JPE$\downarrow$} & \multicolumn{2}{c}{DTW-JPE$\downarrow$}
& \multicolumn{2}{c}{DTW-PA-JPE$\downarrow$} & \multicolumn{2}{c}{DTW-JPE$\downarrow$}
& \multicolumn{2}{c}{DTW-PA-JPE$\downarrow$} & \multicolumn{2}{c}{DTW-JPE$\downarrow$} \\
\cmidrule(lr){2-3}\cmidrule(lr){4-5}
\cmidrule(lr){6-7}\cmidrule(lr){8-9}
\cmidrule(lr){10-11}\cmidrule(lr){12-13}
& Body & Hand & Body & Hand
& Body & Hand & Body & Hand
& Body & Hand & Body & Hand \\
\midrule
NAR$^{\blacklozenge}$ \cite{hwang2021non}
& 13.94 & 11.80 & -- & --
& -- & -- & -- & --
& -- & -- & -- & -- \\
\rowcolor{black!5}Prog. Trans.$^{\Diamond}$~\cite{saunders2020progressive}
& 14.15 & 11.57 & 14.74 & 30.17
& 15.98 & 12.91 & 16.30 & 32.63
& 13.67 & 11.95 & 15.01 & 31.77 \\
Text2Mesh$^{\Diamond}$~\cite{stoll2022there}
& 13.99 & 13.47 & 15.50 & 32.97
& 13.47 & 12.10 & 13.76 & 30.37
& 13.48 & 12.06 & 14.04 & 31.64 \\
\rowcolor{black!5}Adv. Train.$^{\blacklozenge}$~\cite{saunders2020adversarial}
& 13.78 & 11.17 & -- & --
& -- & -- & -- & --
& -- & -- & -- & -- \\
T2S-GPT$^{\Diamond}$~\cite{yin-etal-2024-t2s}
& 11.48 & 6.39 & 12.65 & 18.44
& 11.94 & 5.93 & 12.32 & 15.43
& 10.38 & 6.47 & 11.65 & 19.09 \\
\rowcolor{black!5}NSA~\cite{baltatzis2024neural}
& 7.83 & 7.33 & -- & --
& -- & -- & -- & --
& -- & -- & -- & -- \\
S-MotionGPT~\cite{jiang2023motiongpt}$^{\Diamond\dagger}$
& 11.23 & 4.39 & 12.41 & 13.74
& 10.81 & 3.78 & 11.58 & 11.31
& 9.45 & 3.41 & 10.42 & 9.08 \\
\rowcolor{black!5}SOKE$^{\ddagger}$ \cite{zuo2025soke}
& 6.82 & 2.35 & 7.75 & 10.08
& 6.24 & 1.71 & 7.38 & 9.68
& 4.77 & \textbf{1.38} & 6.04 & \textbf{7.72} \\
\midrule
\paperTitleShort\textsubscript{w/o \ac{slt}}
& 4.52 & 2.42 & 5.02 & \textbf{8.94}
& 4.13 & 1.59 & 4.77 & 9.94
& 3.54 & 1.54 & 4.17 & 9.14 \\
\rowcolor{black!5}
\paperTitleShort\textsubscript{w/ \ac{slt}}
& \textbf{4.28} & \textbf{2.28} & \textbf{4.86} & 9.44
& \textbf{3.97} & \textbf{1.51} & \textbf{4.63} & \textbf{9.28}
& \textbf{3.41} & 1.46 & \textbf{4.02} & 8.55 \\
\bottomrule
\end{tabular}
}
\end{table*}

\subsection{Evaluation on Sign Language Production}\label{sec:slp_results}
\noindent\textbf{Quantitative Comparison:}
Our method achieves state-of-the-art geometric \ac{slp} performance across three benchmarks (\cref{tab:sota_slp}).
We outperform both non-tokenized~\cite{baltatzis2024neural} and tokenized baselines, with substantial gains in body motion across all datasets.
Following the protocol established across prior \ac{slp} work ($\Diamond$ rows reported by SOKE~\cite{zuo2025soke}), each system is evaluated against its own pseudo-GT; a fully controlled comparison would require all methods to train on a common pseudo-GT distribution, which our SMPL-FX release is intended to enable.
Notably, SOKE~\cite{zuo2025soke} incorporates retrieval-augmented generation~\cite{chen2024benchmarking} and benefits from additional isolated sign dictionaries; our method achieves superior results without external retrieval or lexicon supervision while also modeling the full facial expression space.

The improvements vary across datasets.
On How2Sign (\ac{asl}, 30K sentences) and CSL-Daily (\ac{csl}, 20K sentences), body DTW-PA-JPE improves by 37\% and 36\% respectively versus SOKE, with consistent hand improvements.
Phoenix14T (\ac{dgs}, 8K sentences) is the smallest dataset with lower video quality that constrains pseudo-GT extraction fidelity; gains here are accordingly more modest (body 29\% versus SOKE), and the auxiliary translation objective yields smaller back-translation improvements (Phoenix: +4.1\% BLEU-4; CSL-Daily: +22.4\% BLEU-4; \cref{tab:slp_extra}), consistent with the noisier signal available at training time.
Hand metrics on Phoenix14T's noisier fittings marginally trail SOKE (DTW-PA 1.46 vs.\ 1.38) while remaining within the same regime; as the smallest and lowest-resolution (210$\times$260) corpus, Phoenix14T yields the noisiest monocular hand pseudo-GT, and SOKE's retrieval of isolated-sign hand exemplars may help most precisely where data is scarce.. The representational contribution of \paperTitleShort{} is to the face channel, and the body and hand gains elsewhere indicate the unified vocabulary does not sacrifice manual quality.\looseness=-1
Geometric error captures body and hand motion fidelity, but does not reflect the most significant representational gap between systems: the face channel.
No prior 3D \ac{slp} system produces face parameters of comparable richness, so a direct head-to-head face metric is undefined; we report our face DTW-VPE of \textbf{1.85}, \textbf{1.91}, and \textbf{2.23} on How2Sign, CSL-Daily, and Phoenix14T as reference numbers for future comparison, and substantiate the face channel through the qualitative comparison and user study below, the NMFs-CSL probe (\cref{sec:slr_experiments}), and the FSQ-vs-VQ ablation (\cref{sec:ablation_studies}).
We will publicly release our SMPL-FX code and fittings to enable standardized facial evaluation and address the lack of 3D facial fittings in sign data.

\begin{figure*}[htbp]
    \centering
     \includegraphics[trim={5.75cm 0cm 2cm .5cm},clip,width=1.\linewidth]{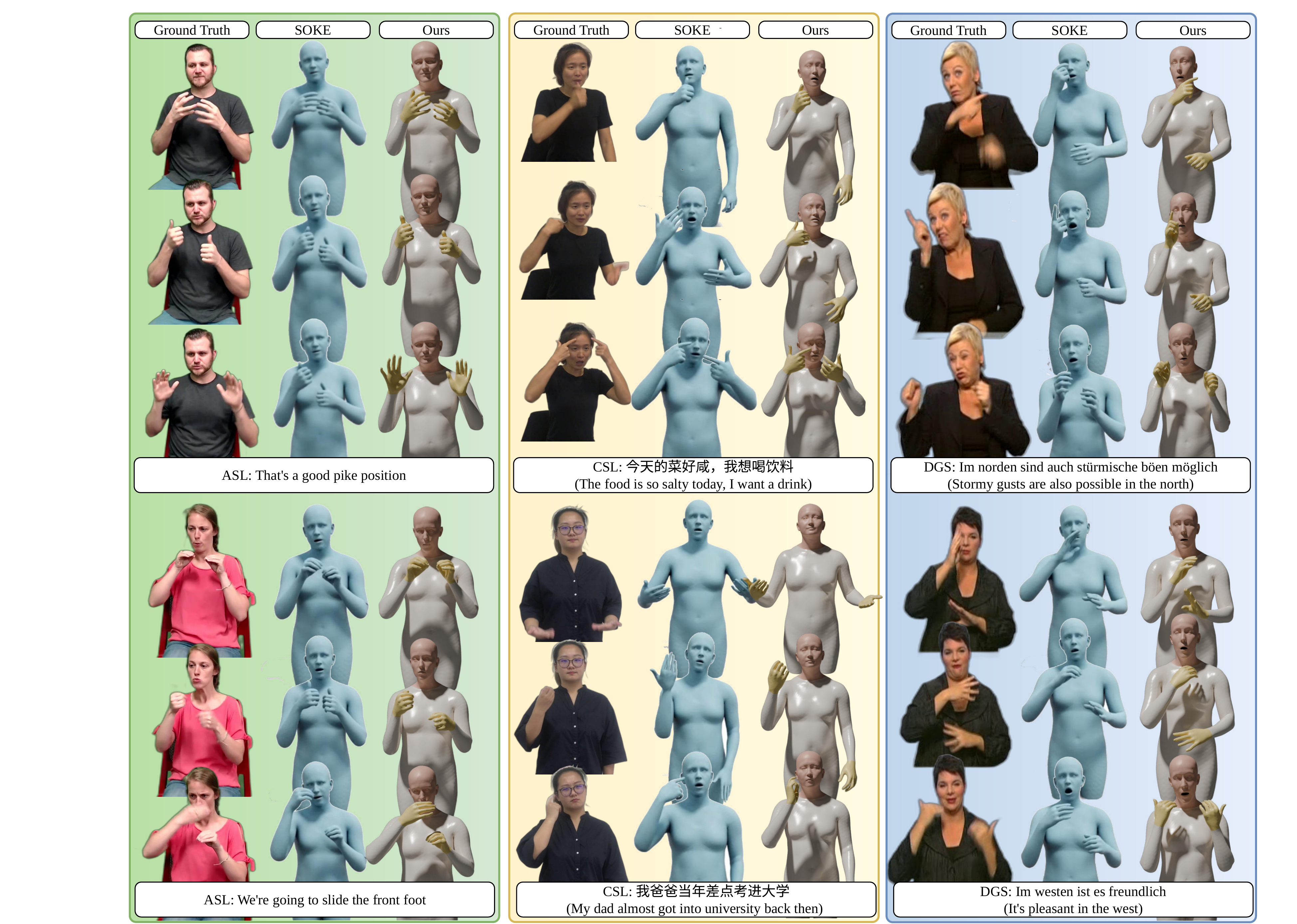}
    \vspace{6pt}
    \caption{\textbf{Qualitative comparisons with SOKE~\cite{zuo2025soke}} over How2Sign (left), CSL-Daily (middle), and Phoenix14T (right). Across all three datasets, SOKE's faces vary mainly in jaw aperture (open vs.\ closed) with minimal additional expression, consistent with the SMPL-X expression range visible in the GT fittings of \cref{fig:smplfx}. \paperTitleShort{}'s SMPL-FX parameterisation allows for \acp{nmf} to be rendered such as mouth shapes and brow movements over the same sequences.}\label{fig:slp_qual}
\end{figure*}

\noindent\textbf{Qualitative Comparison:} SOKE's faces in \cref{fig:slp_qual} vary mainly via jaw movement, as the jaw is a rotation rather than a PCA expression parameter, with limited additional expression.
This is an expected consequence of its ten-dimensional expression fittings, whose limited range is already visible in the GT fittings (\cref{fig:smplfx}).
Additionally, the jaw fitting is not particularly accurate to the GT.
\paperTitleShort{} additionally renders mouth shapes and brow movements via SMPL-FX, recovering more of the articulation channels that \acp{nmf} rely on.

\begin{figure}[htbp]
    \centering
    \includegraphics[width=0.52\linewidth]{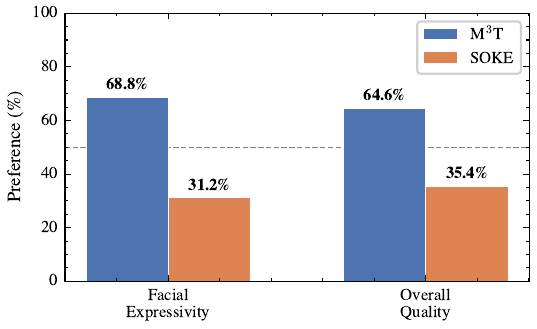}
    \vspace{6pt}
    \caption{\textbf{User study.} Pairwise preference rates (\%) for \paperTitleShort{} vs SOKE~\cite{zuo2025soke} on facial expressivity and overall motion quality. $N=16$ participants, 6 video pairs spanning How2Sign, CSL-Daily, and Phoenix14T.}
    \label{fig:userstudy}
\end{figure}

\noindent\textbf{User Study.} We conduct a pairwise preference study comparing \paperTitleShort{} against SOKE~\cite{zuo2025soke}. Sixteen participants view six randomised video pairs (two per dataset) and select their preferred sequence on two measures: facial expressivity and overall motion quality.
Participants comprise two native Deaf BSL signers and fourteen BSL signers of Level 2--3 proficiency, some with additional proficiency in other sign languages; this rater pool matches or exceeds those of prior \ac{slp} studies~\cite{zuo2025soke,baltatzis2024neural,yin-etal-2024-t2s,wang2025advancedsignlanguagevideo}.
As no participant was fluent in ASL, CSL, or DGS, the protocol assesses perceived naturalness and expressivity rather than comprehensibility or the grammatical correctness of the generated \acp{nmf}; a comprehension study with native signers of each target language remains important future work.
\paperTitleShort{} is preferred in \textbf{68.8\%} of facial-expressivity comparisons and \textbf{64.6\%} of overall-quality comparisons (\cref{fig:userstudy}), with the two native signers' preferences consistent with the aggregate. Lip articulation is identified as the most common residual weakness, consistent with the monocular fitting limitations of our pseudo-GT pipeline.\looseness=-1

While \paperTitleShort~improves facial expressivity, we observe some residual limitations.
Facial outputs occasionally exhibit artifacts under severe occlusion or motion blur.
Furthermore, rapid finger movements remain challenging, particularly on the How2Sign dataset, which contains frequent fingerspelling.
We attribute this primarily to the fixed $4\times$ temporal downsampling in our VAE encoders, rather than a fundamental limitation of the underlying autoregressive modeling.
Finally, we note a tendency for the model to spatially compress the signing space, occasionally regressing hand motions toward the center of the body.
Despite limitations, \paperTitleShort{} remains the first expressive face channel in 3D \ac{slp}.
We encourage readers to view the videos on the project page, as they convey both the \acp{nmf} and motion dynamics more clearly than static frames.

\subsection{Evaluation on Sign Language Recognition} \label{sec:slr_experiments}
\begin{table*}[t]
\centering
\scriptsize
\renewcommand{\arraystretch}{1.1}
\setlength{\tabcolsep}{2.5pt}
\captionsetup{font=small, skip=6pt}

\begin{minipage}[t]{0.48\textwidth}
  \vspace{0pt}
  \centering
  \captionof{table}{\textbf{\ac{slp} Back-Translation Results.} Text metrics are ROUGE-L (R) and BLEU-4 (B4). BT scores on ground truth 3DMM parameters given to highlight upper bound of results.}
\label{tab:slp_extra}
\begin{tabularx}{\linewidth}{l XX XX XX}
\toprule
\multirow{2}{*}{Method} & \multicolumn{2}{c}{\textbf{How2Sign}} & \multicolumn{2}{c}{\textbf{CSL-Daily}} & \multicolumn{2}{c}{\textbf{Phoenix14T}} \\
\cmidrule(lr){2-3} \cmidrule(lr){4-5} \cmidrule(lr){6-7}
& R$\uparrow$ & B4$\uparrow$ & R$\uparrow$ & B4$\uparrow$ & R$\uparrow$ & B4$\uparrow$ \\
\midrule
\rowcolor{black!5}\textbf{GT 3DMM} & 18.68 & 2.04 & 27.14 & 7.11 & 35.83 & 11.84 \\
\midrule
\paperTitleShort\textsubscript{w/o SLT} & 12.32 & 1.58 & 17.27 & 4.74 & 31.14 & 10.16 \\
\rowcolor{black!5}\paperTitleShort\textsubscript{w/ SLT} &\textbf{13.15} & \textbf{1.72} & \textbf{20.23} & \textbf{5.80} & \textbf{32.12} & \textbf{10.58} \\
\bottomrule
\end{tabularx}

  \vspace{0pt}

  \captionof{table}{%
\textbf{Ablation of Quantization Method and Parameters.}
Results are averaged across the sign datasets.
Lower PA-VPE indicates more accurate reconstruction; higher codebook usage reflects more efficient token utilization.}
\label{tab:tokenizer}
\begin{tabularx}{\linewidth}{l cc XXX XXX}
\toprule
\multirow{2}{*}{$Q(\cdot)$} & \multirow{2}{*}{BS} & \multirow{2}{*}{HS} & \multicolumn{3}{c}{\textbf{PA-VPE$\downarrow$}} & \multicolumn{3}{c}{\textbf{Codebook \%$\uparrow$}} \\
\cmidrule(lr){4-6}\cmidrule(lr){7-9}
& & & Body & Hand & Face & Body & Hand & Face \\
\midrule
\rowcolor{gray!10}\multicolumn{9}{c}{\textsc{Quantization Method}} \\
VQ  & 16  & 1024 & 20.67 & 5.61 & 1.35 & 67.3 & 72.5 & 78.8 \\
\rowcolor{black!5} FSQ & 16  & 1024 & \textbf{16.11} & \textbf{4.94} & \textbf{1.18} & \textbf{96.4} & \textbf{98.1} & \textbf{99.0} \\
\midrule
\rowcolor{gray!10}\multicolumn{9}{c}{\textsc{Hyperparameter Search}} \\
FSQ & 128 & 1024 & 24.85 & 6.52 & 1.55 & 84.1 & 86.9 & 89.3 \\
\rowcolor{black!5} FSQ & 64 & 1024 & 21.73 & 5.98 & 1.42 & 89.7 & 92.4 & 94.1 \\
FSQ & 32 & 1024 & 18.92 & 5.37 & 1.30 & 93.2 & 95.7 & 97.0 \\
\rowcolor{black!5} FSQ & 16 & 512 & 17.03 & 5.12 & \textbf{1.08} & 94.9 & 97.2 & \textbf{99.5} \\
FSQ & 16  & 2048 & 16.84 & 4.98 & 1.22 & \textbf{97.1} & \textbf{98.5} & 99.3 \\
\rowcolor{black!5} FSQ & 16  & 1024 & \textbf{16.11} & \textbf{4.94} & 1.18 & 96.4 & 98.1 & 99.0 \\
\bottomrule
\end{tabularx}

\end{minipage}
\hfill
\begin{minipage}[t]{0.48\textwidth}
  \vspace{0pt}
  \centering
  \newcommand{\gr}[1]{\textcolor{black!65}{#1}}%
\captionof{table}{\textbf{\ac{slr} results on NMFs-CSL~\cite{hu2021global}}, separated by RGB and pose modalities.
$\dagger$ denotes pre-training on large sign datasets (e.g., YouTubeASL, BOBSL---10--100$\times$ more data than our setting); these rows (greyed) are not directly comparable to methods trained from scratch, and we bold the best score among comparable (non-$\dagger$) rows.
NLASLR~\cite{zuo2023natural} was trained with both RGB and additional pose input.
\textit{Confusing} and \textit{normal} refer to signs distinguished solely by \acp{nmf} and those otherwise disambiguated, respectively.}
\label{tab:nmf_csl_flat_with_splits}
\begin{tabularx}{\linewidth}{p{1.66cm} XX XX XX}
\toprule
\multirow{2}{*}{Method} & \multicolumn{2}{c}{\textbf{Overall}} & \multicolumn{2}{c}{\textbf{Conf.}} & \multicolumn{2}{c}{\textbf{Norm.}} \\
\cmidrule(lr){2-3} \cmidrule(lr){4-5} \cmidrule(lr){6-7}
& T-1$\uparrow$ & T-5$\uparrow$ & T-1$\uparrow$ & T-5$\uparrow$ & T-1$\uparrow$ & T-5$\uparrow$ \\
\midrule
\rowcolor{gray!10}\multicolumn{7}{c}{\textsc{RGB-based Method}} \\
3D-R50~\cite{qiu2017learning} & 62.1 & 82.9 & 43.1 & 72.4 & 87.4 & 97.0 \\
\rowcolor{black!5} DNF~\cite{cui2019deep} & 55.8 & 82.4 & 51.9 & 71.4 & 86.3 & 97.0 \\
I3D~\cite{carreira2017quo} & 64.4 & 88.0 & 47.3 & 81.8 & 87.1 & 97.3 \\
\rowcolor{black!5} TSM~\cite{lin2019tsm} & 64.5 & 88.7 & 42.9 & 81.0 & 93.3 & 99.0 \\
SlowFast~\cite{feichtenhofer2019slowfast} & 66.3 & 86.6 & 47.0 & 77.4 & 92.0 & 98.9 \\
\rowcolor{black!5} GLE-Net~\cite{hu2021global} & 69.0 & 88.1 & 50.6 & 79.6 & 93.6 & 99.3 \\
HMA~\cite{hu2021hand} & 64.7 & 91.0 & 42.3 & 84.8 & \textbf{94.6} & 99.3 \\
\rowcolor{black!5} \gr{NLASLR~\cite{zuo2023natural}$\dagger$} & \gr{83.1} & \gr{98.3} & \gr{--} & \gr{--} & \gr{--} & \gr{--} \\
\gr{SignRep~\cite{wong2025signrep}$\dagger$} & \gr{84.1} & \gr{98.8} & \gr{--} & \gr{--} & \gr{--} & \gr{--} \\
\midrule
\rowcolor{gray!10}\multicolumn{7}{c}{\textsc{Pose-based Method}} \\
ST-GCN~\cite{yan2018spatial} & 59.9 & 86.8 & 42.2 & 79.4 & 83.4 & 96.7 \\
\rowcolor{black!5} SignBERT~\cite{hu2021signbert} & 67.0 & \textbf{95.3} & 46.4 & 92.1 & 94.5 & 99.6 \\
BEST~\cite{zhao2023best} & 68.5 & 94.4 & 49.0 & 90.3 & \textbf{94.6} & \textbf{99.7} \\
\rowcolor{black!5}\gr{ScalingUp~\cite{zhou2025scaling}$\dagger$} & \gr{80.2} & \gr{97.5} & \gr{67.2} & \gr{95.7} & \gr{97.5} & \gr{99.8} \\
\paperTitleShort & \textbf{74.1} & 95.1 & \textbf{58.3} & \textbf{93.0} & 93.5 & 97.6 \\
\bottomrule
\end{tabularx}

\end{minipage}
\end{table*}

We use NMFs-CSL as a lower-bound probe on the facial linguistic content available to the production decoder: the recognition head consumes only the frozen production encoders' hidden states, so any accuracy attributable to face must already be preserved in the tokens our decoder generates from.
The ``confusing'' subset isolates the face-token contribution since these signs differ only in \acp{nmf}; the ``normal'' subset acts as a control where manual features alone suffice.
On the normal subset, \paperTitleShort~scores 93.5\% Top-1 (\cref{tab:nmf_csl_flat_with_splits}), marginally below BEST~\cite{zhao2023best} at 94.6\%.
This slight variance is expected because these standard signs are fully distinguished by hand shape and trajectory; adding face tokens provides no additional discriminative advantage for this subset and may marginally dilute the manual signal.
On the confusing subset, \paperTitleShort{} achieves 58.3\% against 49.0\% for BEST~\cite{zhao2023best}.
Since body and hand features carry no discriminative signal for confusing signs, the performance gap reflects the face tokens' ability to encode grammatically relevant non-manual distinctions that prior systems often represent too coarsely to recover.
Methods such as ScalingUp~\cite{zhou2025scaling} achieve higher overall accuracy by leveraging large-scale pose pre-training on external sign corpora (e.g., YouTubeASL~\cite{uthus2023youtubeasl}, BOBSL~\cite{albanie2021bbc}) with recognition-optimized architectures.
Against BEST~\cite{zhao2023best}, the strongest baseline without external pre-training, \paperTitleShort{} achieves a +9.3\,pp gain on the confusing subset, confirming that the face tokens encode discriminative \ac{nmf} information available to the production decoder.
As this probe operates on real motion, it validates the token vocabulary rather than the generated output; the production model, however, generates from exactly this vocabulary, and back-translation over the generated sequences (which include the face channel) recovers 81--89\% of the ground-truth ceiling on all three corpora (\cref{tab:slp_extra}).
We do not run the classifier on generated motion, since it is trained on real isolated clips and the resulting scores would mix generation error with domain shift.
Direct evaluation of generated \acp{nmf} by native signers remains future work.

\begin{figure}[htbp]
  \centering
  \begin{subfigure}[b]{0.42\linewidth}
    \includegraphics[width=\linewidth]{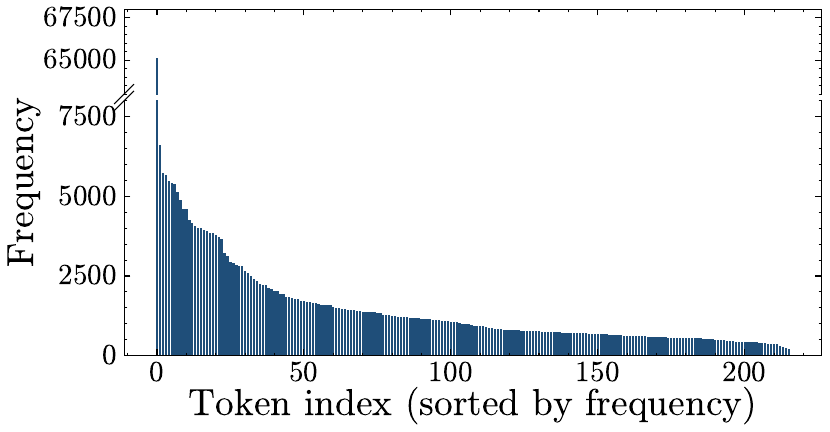}
    \subcaption{FLAME VQ}
  \end{subfigure}
  \hfill
  \begin{subfigure}[b]{0.42\linewidth}
    \includegraphics[width=\linewidth]{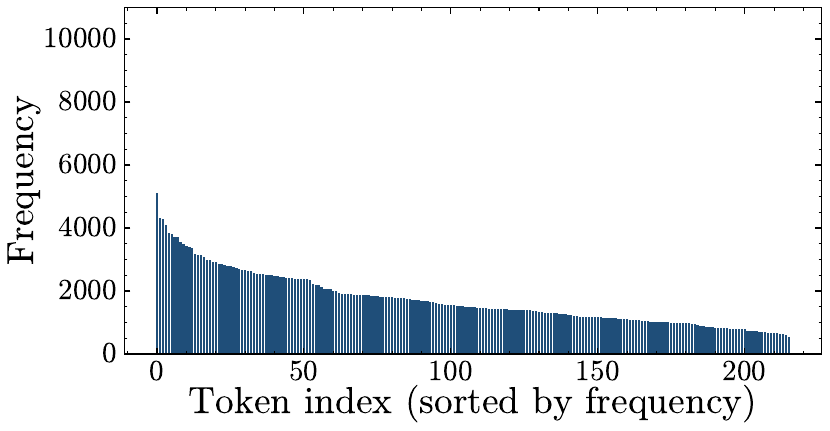}
    \subcaption{FLAME FSQ}
  \end{subfigure}
  \vspace{6pt}
  \caption{\textbf{Face token frequency distributions.} VQ produces a heavily skewed distribution — most face codes are unused, a handful dominate — while FSQ yields near-uniform coverage. Rarely-used tokens receive insufficient gradient signal, undermining the auxiliary translation objective's ability to ground face embeddings semantically; FSQ's structured grid eliminates this bottleneck.}
  \label{fig:token_freq}
\end{figure}

\subsection{Tokenization Ablation}\label{sec:ablation_studies}

We ablate our tokenization framework across quantization methods and architectural hyperparameters, evaluating reconstruction fidelity via PA-VPE and codebook utilization (\cref{tab:tokenizer}).

\textbf{FSQ vs.\ VQ.} The VQ baseline shares our full training recipe (independent per-modality training and the shared mirrored hand encoder), so the differences in \cref{tab:tokenizer} are attributable to the quantizer alone. FSQ improves utilization and reconstruction across all three modalities, reducing average reconstruction error by 14.2\%. Body VQ utilization is lowest in absolute terms (67.3\%) and gains the most in absolute points (+29.1 to 96.4\%), but the recovery to \textbf{99.0\%} face utilization is the consequential one: body and hand parameters are joint rotations in a high-variance pose space, where an unused entry costs accuracy but not expressiveness, whereas most of the face parameters $\boldsymbol{\psi}_f \in \mathbb{R}^{108}$ lie in a more compact PCA expression space in which each entry may map to a discrete articulation (a mouthing, a brow raise), so an unoccupied entry could remove an entire \ac{nmf}. The same property may also explain why the FSQ-trained face tokens suffice to discriminate the NMFs-CSL confusing split (\cref{sec:slr_experiments}). Token frequency distributions in \cref{fig:token_freq} further solidify this asymmetry: VQ concentrates probability mass on a small subset of face codes while FSQ produces near-uniform coverage. Codebook sizes are fixed at $|C_b|{=}100$, $|C_h|{=}150$, and $|C_f|{=}216$ to match prior work~\cite{zuo2025soke}; the face codebook is larger to accommodate FLAME's higher-dimensional expression space.

\textbf{Architectural sensitivity.} Encoder hidden width (HS) trades off facial and body fidelity. HS=512 achieves peak facial reconstruction (1.08 PA-VPE, 99.5\% utilization) but lacks the capacity to model higher-variance body dynamics (17.03 body error). HS=1024 is most robust across the full skeletal structure; HS=2048 yields the highest body utilization (97.1\%) with diminishing face returns. FSQ is also sensitive to batch size, with reconstruction degrading from BS=16 to BS=128, which we attribute to early convergence toward a subset of codes even without a learned codebook.

\section{Conclusion}\label{sec:conclusion}

\acp{nmf} carry grammatical weight that manual articulation alone cannot convey.
\paperTitleShort{} addresses this by pairing a richer face geometry (SMPL-FX) with collapse-free tokenization (FSQ; 99\% face utilization vs.\ 78.8\% for VQ) and an auxiliary translation objective that grounds the learned tokens in meaning, further improving production quality (+3.8\% JPE, +11.8\% back-translation BLEU-4, +9.0\% ROUGE-L).
Trained on this multi-modal vocabulary, our autoregressive model achieves state-of-the-art results across three sign datasets without retrieval augmentation or external dictionaries.\looseness=-1

Several limitations remain: reconstruction quality degrades under occlusion and low-quality footage, most visibly on Phoenix14T; fingerspelling remains challenging under 4$\times$ temporal compression, a limitation shared by existing 3D \ac{slp} methods; and back-translation and joint position error remain imperfect proxies for sign quality.\looseness=-1

Overall, our results indicate that the primary barrier to producing \acp{nmf} in 3D \ac{slp} has been representational rather than architectural.
These improvements concern the token vocabulary rather than the decoder, transferring to concurrent token-based sign generators such as MaDiS~\cite{zuo2026madis}, and we release the SMPL-FX code and fittings as a shared resource.\looseness=-1

\section*{Acknowledgements}
This work was supported by EPSRC grant APP24554 (SignGPT--EP/Z535370/1), EPSRC grant APP78083 (UMCS UKRI3927) and through funding from Google.org via the AI for Global Goals scheme.
The authors acknowledge the use of the Isambard-AI National AI Research Resource (AIRR) funded by UK DSIT via UKRI and STFC [ST/AIRR/I-A-I/1023].
This work reflects only the authors' views and the funders are not responsible for any use that may be made of the information it contains.

\bibliography{main}

\clearpage
\appendix
\renewcommand{\thefigure}{A.\arabic{figure}}
\setcounter{figure}{0}
\renewcommand{\thetable}{A.\arabic{table}}
\setcounter{table}{0}
\renewcommand{\theequation}{A.\arabic{equation}}
\setcounter{equation}{0}

\begin{figure*}[!htb]
    \centering
    \includegraphics[width=0.85\textwidth]{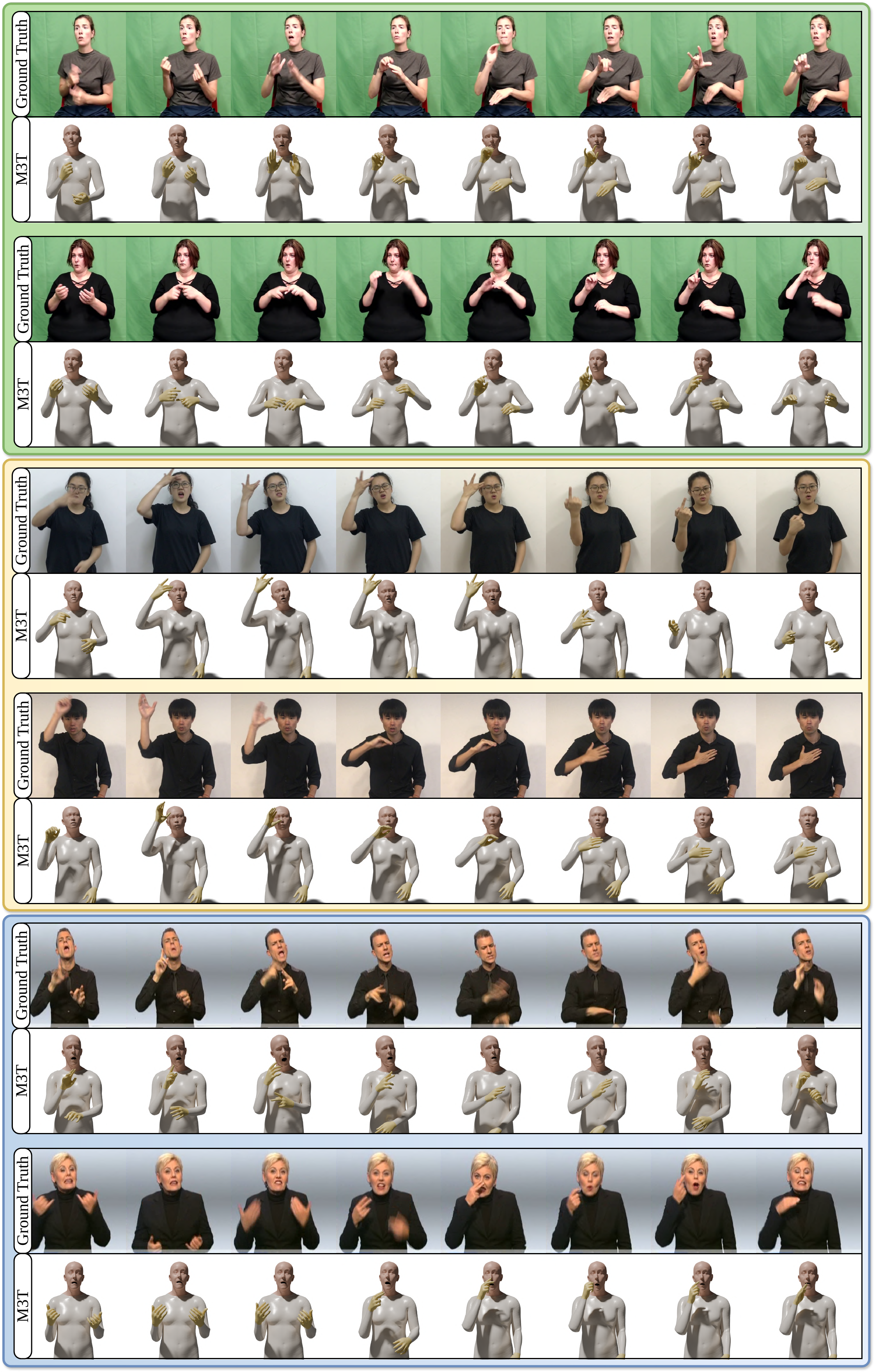}
    \vspace{6pt}
    \caption{\textbf{Additional Qualitative Results.} Extended sequences from \paperTitleShort{} on How2Sign (top), CSL-Daily (middle), and Phoenix14T (bottom). Note the variation in facial expressions across frames, with mouth movements and eyebrow raises coordinated with the manual signs. Body posture and hand configurations maintain temporal coherence over these longer sequences.} \label{supp:fig:extra_qual}
\end{figure*}

This appendix provides additional technical details, analyses, and visualizations that complement the main text.
\Cref{supp:qual} includes extended qualitative results over longer sequences.
\Cref{supp:smplfx} expands on the foundations of SMPL-FX with a detailed formulation of the hybrid body--face representation.
\Cref{supp:tokviz} analyses the motion vocabulary statistics across modalities.
\Cref{supp:imp} presents comprehensive implementation and training details to facilitate reproducibility.
Videos of the qualitative results shown in \Cref{fig:slp_qual} and \Cref{supp:fig:extra_qual} are available on the project page.

\section{Additional Qualitative Results} \label{supp:qual}

The extended qualitative examples (\cref{supp:fig:extra_qual}) show \paperTitleShort{}'s expressive and accurate signing over longer sequences.
We encourage the reader to view the accompanying videos, which clearly convey the fluidity and expressivity of the generated motion.
The user study (\cref{sec:slp_results}) additionally noted that generated motion tends to occupy a smaller signing space than real signing, likely due to temporal compression in the tokenization stage.

\section{SMPL-FX} \label{supp:smplfx}
This section details the SMPL-FX body model summarized in \cref{sec:3dmm_prep}.
We first review the shared \ac{3dmm} formulation before describing the FLAME integration and per-frame fitting pipeline.
\Cref{fig:smplfx} compares expressions reconstructed with baseline SMPL-X parameters from~\cite{zuo2025soke} against our SMPL-FX fitting, showing that SMPL-FX preserves NMFs that SMPL-X cannot.

\noindent\textbf{3DMM Preliminaries.}
Parametric models such as SMPL~\cite{loper2015smpl}, MANO~\cite{mano}, and FLAME~\cite{FLAME} are widely used \acp{3dmm}~\cite{blanzvetter3dmm} that share a mesh generation formulation:
\begin{equation}\label{supp:eq:3dmm_full}
\mathbf{M}(\boldsymbol{\beta}, \boldsymbol{\theta}, \boldsymbol{\psi}) = W\big(T_P(\boldsymbol{\beta}, \boldsymbol{\theta}, \boldsymbol{\psi}), J(\boldsymbol{\beta}), \boldsymbol{\theta}, \mathbf{W}\big),
\end{equation}
where $W(\cdot)$ applies linear blend skinning to a template mesh $\bar{\mathbf{T}}$ using learned blend weights $\mathbf{W}$ and joint locations $J(\boldsymbol{\beta})$.
The rest-pose mesh is given by:
\begin{equation}\label{supp:eq:3dmm_neutral}
T_P(\boldsymbol{\beta}, \boldsymbol{\theta}, \boldsymbol{\psi}) = \bar{\mathbf{T}} + B_S(\boldsymbol{\beta}) + B_P(\boldsymbol{\theta}) + B_E(\boldsymbol{\psi}),
\end{equation}
where $B_S$, $B_P$, and $B_E$ model identity, articulation, and expression.
Only FLAME and SMPL-X~\cite{pavlakos2019expressive} include expression blendshapes $B_E$.
\Cref{supp:tab:3dmm} compares mesh vertex counts and parameter spaces across all relevant models.

\noindent\textbf{Formulation.}
SMPL-FX inherits from SMPL-X and replaces its head region with a FLAME mesh.
A precomputed vertex correspondence map identifies which SMPL-X vertices correspond to FLAME vertices.
At initialization, the SMPL-X template vertices at those indices are replaced with FLAME's template geometry, offset by the mean displacement to preserve global alignment.
FLAME's shape and expression blendshape directions are then retargeted into full-body buffers: each buffer has the dimensionality of the SMPL-X mesh but is non-zero only at the FLAME-corresponding indices.
SMPL-X's own shape blendshapes are zeroed at the same indices so that body shape does not deform the face.
To avoid a visible seam at the neck boundary, vertex blendshape weights are linearly interpolated across concentric neck rings defined by FLAME's topology: ratios decay from 0.9 (outermost ring, closest to the body) to 0.1 (innermost, closest to the jaw), smoothly transitioning between SMPL-X and FLAME influence.
During the forward pass, FLAME shape and expression blendshapes are applied first, followed by optional eyelid displacements~\cite{giebenhain2025pixel3dmm} (two per-vertex offset buffers scaled by scalar eyelid parameters).
SMPL-X body-shape blendshapes (with the face zeroed out) are then added, and standard linear blend skinning is applied using SMPL-X's joint regressor and blend weights.
Teeth are added as rigid geometry attached to the jaw and lower-jaw joints, inheriting their transformations from the skinning step~\cite{qian2024gaussianavatars}.
The final SMPL-FX mesh comprises 10{,}995 vertices and 42 pose joints (10 body + 30 hands + 2 FLAME eyes), after discarding 11 SMPL-X lower-body joints unused in sign production and re-representing the jaw rotation within the 108-dimensional expression vector.
The pose space comprises hand poses $\boldsymbol{\theta}_h^l, \boldsymbol{\theta}_h^r \in \mathbb{R}^{15 \times 6}$ (15 joints each) and body pose $\boldsymbol{\theta}_b \in \mathbb{R}^{10 \times 6}$, both in the 6D continuous rotation representation~\cite{zhou2019continuity}. The identity blendshape $\beta \in \mathbb{R}^{310}$ combines SMPL-X's 10 body shape coefficients with FLAME's 300 head shape coefficients.
The expression space is 108-dimensional: 100 FLAME expression coefficients, 2 eyelid parameters, and a 6D continuous representation of the jaw pose.

\noindent\textbf{Extraction Pipeline.}
Fitting all SMPL-FX parameters jointly from monocular video is difficult because expression blendshapes produce small vertex displacements that are easily dominated by body pose.
Each modality is therefore extracted with a dedicated tracker---NLF~\cite{sarandi2024neural_nlf} for body, WiLoR~\cite{potamias2025wilor} for hands, and Pixel3DMM~\cite{giebenhain2025pixel3dmm} for face---then fused and refined via joint optimization.
The loss combines three terms:
\begin{equation}\label{supp:eq:smplfx_opt}
\mathcal{L}_{\text{fit}} = \lambda_{\text{kp}} \mathcal{L}_{\text{kp}} + \lambda_{\text{acc}} \mathcal{L}_{\text{acc}} + \lambda_{\text{reg}} \mathcal{L}_{\text{reg}},
\end{equation}
where $\mathcal{L}_{\text{kp}}$ is an L2 reprojection loss against 2D keypoints from RTMPose~\cite{jiang2023rtmose,mmpose2020} ($\lambda_{\text{kp}}=1.0$),
$\mathcal{L}_{\text{acc}}$ penalizes vertex acceleration for temporal smoothness ($\lambda_{\text{acc}}=0.5$),
and $\mathcal{L}_{\text{reg}}$ regularizes pose and expression parameters toward the per-tracker initialization ($\lambda_{\text{reg}}=0.01$).
Optimization is performed per clip using Adam on an RTX 3090.
While SMPL-FX substantially increases facial expressivity over SMPL-X, mesh quality is ultimately bounded by the upstream monocular fitting stage, where lip detail is particularly difficult to recover from a single viewpoint.
This limitation propagates through the full pipeline, since the FSQ-VAE can only tokenize what the fitting provides and the production model can only generate what the tokenizer encodes.
Feedback from the user study in \cref{sec:slp_results} corroborates this, with participants identifying lip articulation as a common weakness.

\begin{table*}[htbp]
\centering
\small
\renewcommand{\arraystretch}{1.1}
\captionsetup{font=small, skip=6pt}
\begin{minipage}[t]{0.46\textwidth}
  \vspace{0pt}
  \centering
  \captionof{table}{\textbf{Mesh resolution and parameter dimensions.} SMPL-FX enhances SMPL-X's head with FLAME's expression space, removing joints irrelevant to signing.} \label{supp:tab:3dmm}
  \begin{tabular}{lcccc}
  \toprule
  \multirow{2}{*}{\textbf{Model}} & \multirow{2}{*}{\textbf{Vertices}} & \multicolumn{3}{c}{\textbf{Parameters}} \\
  \cmidrule(lr){3-5}
   &  & \textbf{Shape} & \textbf{Pose} & \textbf{Expr.} \\
  \midrule
  \rowcolor{black!5}SMPL~\cite{loper2015smpl}    & 6{,}890   & 10   & 21  & -- \\
  MANO~\cite{mano}             & 778      & 10   & 15  & -- \\
  \rowcolor{black!5}FLAME~\cite{FLAME}           & 5{,}023   & 300  & 4   & 100 \\
  SMPL-X~\cite{pavlakos2019expressive} & 10{,}475  & 10   & 54  & 10 \\
  \rowcolor{black!5}\emph{SMPL-FX}      & 10{,}995  & 310  & 42  & 108 \\
  \bottomrule
  \end{tabular}
\end{minipage}
\hfill
\begin{minipage}[t]{0.46\textwidth}
  \vspace{0pt}
  \centering
  \captionof{table}{\textbf{Standard deviation of token counts across SMPL-FX modalities.} FSQ yields $\approx82\%$ lower SD than VQ, indicating more uniform codebook utilization.} \label{supp:tab:token}
  \begin{tabular}{lccc}
  \toprule
  \textbf{} & \textbf{SMPL} & \textbf{MANO} & \textbf{FLAME} \\
  \midrule
  \rowcolor{black!5}\textbf{VQ SD}  & 6{,}964  & 9{,}658  & 4{,}499 \\
  \textbf{FSQ SD} & 1{,}316\textsubscript{\textcolor{softpink}{$\downarrow$81\%}} &
                    1{,}605\textsubscript{\textcolor{softpink}{$\downarrow$83\%}} &
                    842\textsubscript{\textcolor{softpink}{$\downarrow$81\%}} \\
  \bottomrule
  \end{tabular}
\end{minipage}
\end{table*}

\section{Statistics on Sign Motion Vocabulary} \label{supp:tokviz}

\begin{figure*}[h]
\centering
\begin{subfigure}[t]{0.48\textwidth}
    \subcaption{SMPL VQ Token Frequencies}

    \includegraphics[width=\linewidth]{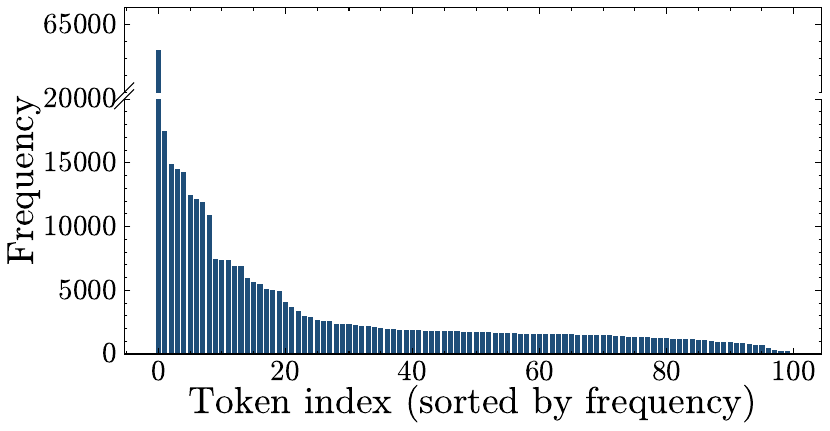}
\end{subfigure}
\hfill
\begin{subfigure}[t]{0.48\textwidth}
    \subcaption{SMPL FSQ Token Frequencies}

    \includegraphics[width=\linewidth]{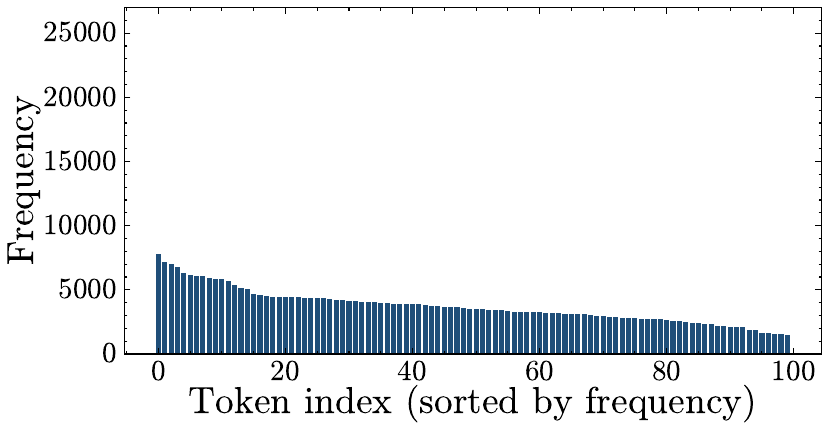}
\end{subfigure}

\begin{subfigure}[t]{0.48\textwidth}
    \subcaption{MANO VQ Token Frequencies}

    \includegraphics[width=\linewidth]{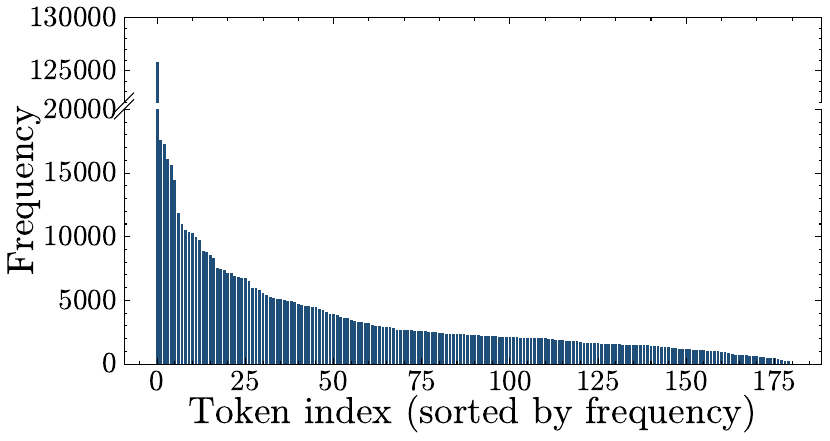}
\end{subfigure}
\hfill
\begin{subfigure}[t]{0.48\textwidth}
    \subcaption{MANO FSQ Token Frequencies}

    \includegraphics[width=\linewidth]{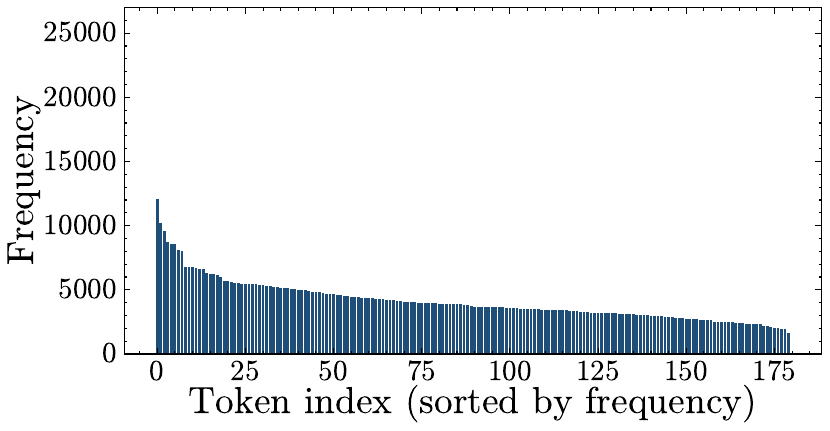}
\end{subfigure}

\begin{subfigure}[t]{0.48\textwidth}
    \subcaption{FLAME VQ Token Frequencies}

    \includegraphics[width=\linewidth]{figs/supp/flame_vq_token_frequency.pdf}
\end{subfigure}
\hfill
\begin{subfigure}[t]{0.48\textwidth}
    \subcaption{FLAME FSQ Token Frequencies}

    \includegraphics[width=\linewidth]{figs/supp/flame_fsq_token_frequency.pdf}
\end{subfigure}

\vspace{6pt}
\caption{
\textbf{Token frequency distributions for body (SMPL), hand (MANO), and face (FLAME) modalities.}
VQ produces skewed distributions; FSQ yields more uniform coverage across all modalities.
The FLAME distributions are also shown in \cref{fig:token_freq} alongside the ablation discussion.
} \label{supp:fig:token_viz}

\end{figure*}

We visualize the token frequency distribution across all test sets in \cref{supp:fig:token_viz}.
For comparison, we include the distribution obtained from a VQ-based VAE with an identical codebook size.
Unlike VQ, which often exhibits pronounced frequency spikes for individual tokens across multiple configurations, FSQ produces a more balanced utilization of the codebook.
This uniformity is evident in the visualization and further supported by the modality-specific standard deviations reported in \Cref{supp:tab:token}.

\section{Implementation and Training Details} \label{supp:imp}

The following sections detail the architecture and training procedures for all components: the FSQ-VAEs (\cref{supp:token}), the \acl{smlm} (\cref{supp:smlm}), the \acl{slr} model (\cref{supp:slr}), and the \acl{bt} models (\cref{supp:bt}).

\subsection{Multi-Modal Sign Motion Tokenization} \label{supp:token}

Our per-modality motion \ac{fsq}-\acp{vae} build upon the \acp{vae} introduced in MotionGPT~\cite{jiang2023motiongpt}.
As noted in \Cref{tab:tokenizer}, certain configurations yield marginally improved reconstructions for specific modalities; however, a shared encoder architecture is used across all three.
Each encoder $\mathcal{E}_m(\cdot)$ (\cref{eq:enc}) begins with a Conv1D layer mapping inputs to an internal width of 1024, followed by 2 downsampling stages (Conv1D with stride 2), each paired with a ResNet-1D stack of six residual blocks.
These blocks employ dilated convolutions with a dilation growth rate of 3.
A final Conv1D reduces channels to 3 for the latent representation.
The decoders $\mathcal{D}_m(\cdot)$ (\cref{eq:dec}) mirror this structure with reversed dilation and nearest-neighbor upsampling replacing strided convolutions.
For quantization, we adopt FSQ~\cite{mentzer2022finite}, which introduces no learnable parameters and is configured via the predefined number of levels per dimension.
The codebook size is given by
$C_m = \prod_{i=1}^{d} L_m^i$,
where $L_m^i$ denotes the number of levels for dimension $i$ in modality $m$.
We select $L$ per modality to achieve codebook sizes comparable to SOKE~\cite{zuo2025soke}, resulting in $L_b = \{5,5,4\}$, $L_h = \{6,5,5\}$, and $L_f = \{6,6,6\}$ for body, hands, and face, respectively.
Each FSQ-VAE comprises approximately 120M parameters, split nearly evenly between encoder and decoder.

\noindent\textbf{Training Setup.} Training is performed on a single NVIDIA RTX 3090 GPU using the Adam optimizer~\cite{adam} with a learning rate of $10^{-4}$ for 100 epochs and a batch size of 8.
We employ cosine annealing with a 25-epoch warm-up and a minimum learning rate of $10^{-6}$. Training uses an L2 reconstruction loss on SMPL-FX parameters and we select the checkpoint with the lowest validation vertex reconstruction error per modality.

\subsection{Sign Motion Language Model} \label{supp:smlm}

The \ac{smlm} is built on mBART-large-cc25~\cite{liu2020multilingual}, following the architecture of MotionGPT~\cite{jiang2023motiongpt} and SOKE~\cite{zuo2025soke}.
The mBART encoder–decoder remains unchanged, preserving its 12 layers and 1024-dimensional hidden size.
To handle multimodal motion data, we extend mBART’s tokenizer (\cref{eq:vocab}) with a motion-specific vocabulary and introduce 5 modality-specific token prediction heads: text, body, left hand, right hand, and face.
Each head is implemented as a linear layer mapping the shared hidden representation to the respective vocabulary indices.
During input token fusion, modality embeddings are combined via equal-weight averaging.
The complete model comprises 396M parameters.

\noindent\textbf{Training Setup.} Our \ac{smlm} is trained on two NVIDIA RTX 3090 GPUs using the AdamW optimizer~\cite{adamw}. A cosine annealing learning rate schedule is employed, starting at $2\times10^{-4}$ and decaying to $10^{-6}$.
Training spans 200 epochs at a batch size of 32, completing within 4 days.
The objective is cross-entropy loss, and the best checkpoint is selected based on mean DTW-MJPE performance on the validation set of all datasets.

\subsection{Sign Language Recognition} \label{supp:slr}
Our \ac{slr} module leverages the pose encoder's intermediate representations, specifically the 1024-dimensional hidden states extracted prior to quantization.
These per-modality features are interleaved into a structured sequence $\{\texttt{body}, \texttt{hand\_l}, \texttt{hand\_r}, \texttt{face}, ...\}$ which is processed by a 3-layer Transformer encoder with a hidden size of 1024 and 4 attention heads.
The output is passed through an MLP classifier. The additional recognition stack introduces 26M parameters on top of the frozen encoder from the tokenizer described in \cref{supp:token}.

\noindent\textbf{Training Setup.} The \ac{slr} model is trained on a single NVIDIA RTX 3090 GPU using the Adam optimizer~\cite{adam}. We employ a cosine annealing schedule starting at $2\times10^{-4}$ and decaying to $10^{-6}$. Training runs for 100 epochs with a batch size of 64, completing in 12 hours.
The loss function is cross-entropy with label-smoothing ($\alpha=0.1$).
The final checkpoint is selected based on the best top-1 accuracy across the validation set.

\subsection{Back-Translation Model} \label{supp:bt}
Our back-translation model builds upon the publicly available Sign Language Transformer (SL-T)~\cite{camgoz2020sign}, following the progressive training approach of~\cite{saunders2020progressive}.
We adapt the base implementation to accommodate SMPL-FX input modalities, concatenating all features into a 360-dimensional input vector.
The original CNN-based feature extractor is removed, allowing the model to directly process 6D pose and facial expressions.

Because SL-T was originally designed exclusively for Phoenix14T~\cite{Camgoz_2018_CVPR_pheonix14}, which is German and includes gloss\footnote{A gloss is a written approximation of a sign's meaning.} annotations, modifications were required for other datasets.
For CSL-Daily, which contains Chinese text, SL-T's character- or word-level tokenizer is incompatible; we address this by employing Jieba~\cite{jieba} for segmentation.
For How2Sign~\cite{duarte2021how2sign}, which lacks gloss annotations, we generate pseudo-glosses\footnote{A pseudo-gloss is an automatically generated approximation of a sign's meaning, often derived from the written translation.} based on part-of-speech tags using the method introduced in Sign2GPT~\cite{wong2024sign2gpt}.

\noindent\textbf{Training Setup.} Models are trained separately for each sign language using an NVIDIA RTX 3090 GPU.
We employ the Adam optimizer~\cite{adam} with a learning rate of $2.5\times10^{-4}$ and a batch size of 256 (128 for How2Sign due to longer sequences).
A plateau-based scheduler reduces the learning rate when validation performance stagnates, with a patience of 10 and a decay factor of $0.8$.
Training completes in under one day for Phoenix14T and CSL-Daily, while How2Sign requires more than twice the time due to increased sequence lengths.
We select the checkpoint achieving the highest validation BLEU-4 score and use this model for back-translation inference to evaluate the \ac{smlm}.

\end{document}